\documentclass{article}
\pdfoutput=1
\usepackage{spconf,amsmath,graphicx,enumitem}
\usepackage{times}
\usepackage{epsfig}
\usepackage{amssymb}
\usepackage{url}
\usepackage{subcaption}
\usepackage{multirow}
\usepackage[multiple]{footmisc}
\usepackage{tabularx,ragged2e,booktabs}
\usepackage{pbox}
\usepackage{xcolor}

\title{Seeing Through Noise:\\
Visually Driven Speaker Separation and Enhancement}

\name{Aviv Gabbay \qquad Ariel Ephrat \qquad Tavi Halperin \qquad Shmuel Peleg}
\address{The Hebrew University of Jerusalem\\
Jerusalem, Israel}

\begin{document}
\maketitle

\begin{abstract}

Isolating the voice of a specific person while filtering out other voices or background noises is challenging when video is shot in noisy environments.
We propose audio-visual methods to isolate the voice of a single speaker and eliminate unrelated sounds. First, face motions captured in the video are used to estimate the speaker's voice, by passing the silent video frames through a video-to-speech neural network-based model. Then the speech predictions are applied as a filter on the noisy input audio. This approach avoids using mixtures of sounds in the learning process, as the number of such possible mixtures is huge, and would inevitably bias the trained model. We evaluate our method on two audio-visual datasets, GRID and TCD-TIMIT, and show that our method attains significant SDR and PESQ improvements over the raw video-to-speech predictions, and a well-known audio-only method.
\end{abstract}


\section{Introduction}
\label{sec:intro}
Single channel speaker separation and speech enhancement have been extensively researched \cite{Bronkhorst2000, ephraim1984speech}. Neural networks have recently been trained to separate audio mixtures into their sources \cite{chen2017single}. These models were able to learn unique speech characteristics as spectral bands, pitches and chirps \cite{isik2016single}. The main difficulty of audio-only approaches is their poor performance in separating similar human voices, such as same-gender mixtures.

We first describe the separation of a mixed speech of two speakers whose faces are visible in the video. We continue with the isolation of the speech of a single visible speaker from background sounds.
This work builds upon recent advances in machine speechreading, generating speech from visible motion of the face and mouth \cite{vid2speech, ephrat2017improved, cornu2017}. 

Unlike other methods which utilize models trained on mixtures of speech and noise or two voices, our approach is speaker dependent and noise-invariant. This allows us to train models using far less data, and still obtain good results, even in cases of two overlapping voices of the same person.

\subsection{Related work}
\label{ssec:related}
\paragraph*{Audio-only speech enhancement and separation}
Previous methods for single-channel, or monaural, speech enhancement and separation mostly use audio only input. The common \emph{spectrographic masking} approach generates masking matrices containing time-frequency (TF) components dominated by each speaker \cite{reddy2007soft,jin2009supervised}. Huang \emph{at al.} \cite{huang2014deep} are among the first to use a deep learning-based approach for speaker dependent speech separation.

Isik \emph{et al.} \cite{isik2016single} tackle the single-channel multi-speaker separation using \emph{deep clustering}, in which discriminatively-trained speech embeddings are used as the basis for clustering and separating speech. Kolbaek \emph{et al.} \cite{pit_kolbaek} introduce a simpler approach in which they use a permutation-invariant loss function which helps the underlying neural network discriminate between the different speakers.

\vspace{-1.2em}
\paragraph*{Audio-visual speech processing}
Recent research in audio-visual speech processing makes extensive use of neural networks. The work of Ngiam \emph{et al.} \cite{ngiam2011multimodal} is a seminal work in this area. Neural networks with visual input have been used for lipreading \cite{chung2016lip}, sound prediction \cite{owens2016visually} and for learning unsupervised sound representations \cite{aytar2016soundnet}.

Work has also been done on audio-visual speech enhancement and separation \cite{Girin2001AudiovisualEO,wang2005video}. Kahn and Milner \cite{khan2013speaker,khan2016audio} use hand-crafted visual features to derive binary and soft masks for speaker separation. Hou \emph{et al.} \cite{hou2017audio} propose CNN based models to enhance noisy speech. Their network generates a spectrogram representing the enhanced speech.

\vspace{-0.5em}
\section{Visually-derived Speech Generation}
\label{sec:video-to-speech}
Several approaches exist for generation of intelligible speech from silent video frames of a person speaking \cite{vid2speech, ephrat2017improved, cornu2017}. In this work we rely on $vid2speech$ \cite{ephrat2017improved}, briefly described in Sec.~\ref{ssec:vid2speech}. It should be noted that these methods are \emph{speaker dependent}, meaning a separate, dedicated model must be trained per speaker.

\subsection{Vid2speech}
\label{ssec:vid2speech}
In a recent paper, Ephrat \emph{et al.} \cite{ephrat2017improved} present a neural network-based method for generating speech spectrograms from a sequence of silent video frames of a speaking person. Their model takes two inputs: $(i)$ a video clip of $K$ consecutive frames, and $(ii)$ a ``clip'' of $(K\!-\!1)$ dense optical flow fields between consecutive frames.
The network architecture consists of a dual-tower ResNet \cite{he2016deep} which takes the aforementioned inputs and encodes them into a latent vector representing the visual features, which is subsequently fed into a series of two fully connected layers, generating mel-scale spectrogram predictions. This is followed by a post-processing network which aggregates multiple consecutive predictions and maps them to a linear-scale spectrogram representing the final speech prediction.

\section{Audio-Visual Speech Separation}
\label{sec:approach}
We propose to examine the spectrogram of the audio input, a mixture of multiple sources, and to assign each time-frequency (TF) element to its respective source. 
The generated spectrograms are used to reconstruct the estimated individual source signals.

The above assignment operation is based on the estimated speech spectrogram of each speaker, as generated by a video-to-speech model from Sec.~\ref{sec:video-to-speech}. Since the video-to-speech process does not generate perfect speech signals, we use them only as prior knowledge for separating the noisy mixture.

\vspace{-1em}
\subsection{Speech separation of two speakers}
\label{ssec:separation}

In this case, two speakers ($D_1$, $D_2$) face a camera using a single microphone. We assume that the speakers are known, i.e. we train two separate video-to-speech networks ($N_1$, $N_2$) in advance, one for each speaker, where $N_1$ is trained using the audio-visual dataset of speaker $D_1$, and $N_2$ is trained on speaker $D_2$.

Given the video of speakers $D_1$ and $D_2$, whose sound track includes their mixed voices, the voice separation process is as follows:

\begin{enumerate}[leftmargin=*,topsep=0pt]
\item 
The faces of speakers $D_1$ and $D_2$ are detected in the video using a face detection method \cite{viola2004robust}.
\vspace{-0.6em}
\item
Speech mel-scale spectrograms $S_1$ and $S_2$ of speakers $D_1$ and $D_2$ are predicted from the respective faces using networks $N_1$ and $N_2$.
\vspace{-0.6em}
\item
A mixture mel-scale spectrogram $C$ is generated from the input audio.
\vspace{-0.6em}
\item
For each $(t,f)$,
\vspace{-1em}
\begin{equation}
\vspace{-0.5em}
F_1(t,f) =
\left\{
	\begin{array}{ll}
		1  & S_1(t,f) > S_2(t,f) \\
		0 & otherwise
	\end{array}
\right.
\end{equation}
\begin{equation}
F_2(t,f) = 1 - F_1(t,f)
\vspace{-1.5em}
\end{equation}
\vspace{-0.6em}
\item
Separated spectrograms $P_i$ for each speaker are generated from the mixture spectrogram $C$ by $P_i = C \odot F_i$, where $\odot$ denotes element-wise multiplication.
\vspace{-0.em}


\item
Separated speech signals are reconstructed from the spectrograms ($P_1$ or $P_2$), preserving the original phase of each isolated frequency.
\end{enumerate}
\vspace{-0.6em}

The binary separation in Step 4 above, where ``winner takes all'', can be modified to generate a ratio mask, which gives each TF bin a continuous value between $0$ and $1$, i.e. the generation of the two masks $F_1$ and $F_2$ can be done by:
\vspace{-1em}
\begin{equation}
F_i(t, f) = \left( \frac{S_i^2(t,f)}{S_1^2(t,f) + S_2^2(t,f)}\right)^\frac{1}{2}, \;\;  i=1,2
\end{equation}


\subsection{Speech enhancement of a single speaker}
\label{ssec:enhancement}
In the speech enhancement case one speaker ($D$) is facing a camera having a single microphone. Background noise, that may include voices of other (unseen) speakers, is also recorded. The task is to separate the speaker's voice from the background noise. As before, we assume that we train in advance a video-to-speech network ($N$) on an audio-visual dataset of this speaker. But unlike speech separation, only a single speech prediction is available.

\begin{figure}[tb]
\centering
\includegraphics[width=1\linewidth]{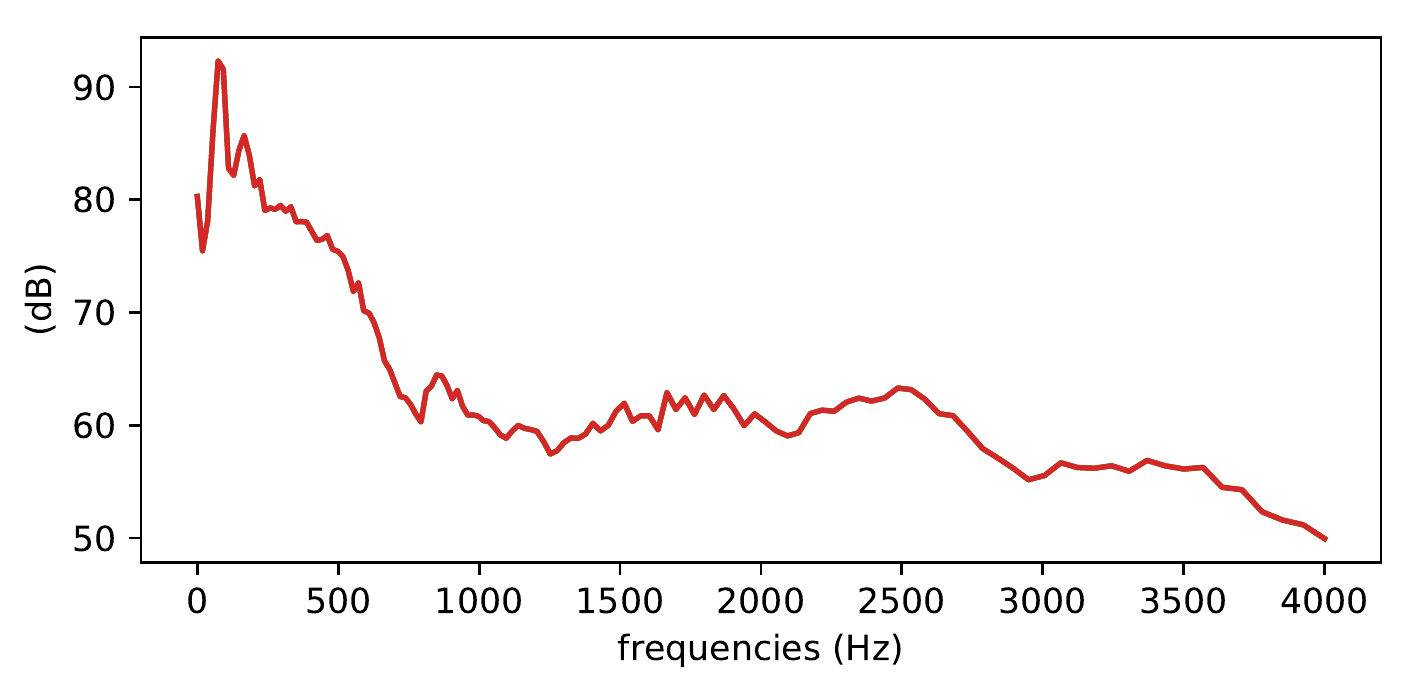}
\caption{Example of a thresholding function based on the Long-Term Speech Spectra (LTSS) of a male speaker. Here, for each frequency $f$, the threshold $\tau(f)$ is set to the 75\% percentile of all seen magnitudes of $f$ in the training data.}
\label{fig:enhancement_thresholds}
\end{figure}

As we assume that the speaker is previously known, we compute the Long-Term Speech Spectra (LTSS) from the speaker's training data, obtaining the distribution of each frequency in the speaker's voice. For each frequency $f$ we pick a threshold $\tau(f)$, indicating when the frequency might come from this speaker's speech, and should be preserved when suppressing the noise. For example, the threshold for a given frequency can be set to the top $X$ percentile (In this case $X$ is a hyperparameter). An example of a thresholding function can be seen in Fig.~\ref{fig:enhancement_thresholds}.

Given a new video of same speaker, having a noisy sound track, the process to isolate the speaker's voice is as follows:
\vspace{-0.4em}
\begin{enumerate}[leftmargin=*]
\item
The thresholding function $\tau(f)$ is computed from the Long-Term Speech Spectra (LTSS) of the training data.
\vspace{-0.6em}
\item
The face of speaker $D$ is detected in the input video using a face detection method.
\vspace{-0.6em}
\item
Speech mel-scale spectrogram $S$ of speaker $D$ is predicted from the detected face using network $N$.
\vspace{-0.6em}
\item
The noisy mel-scale spectrogram $C$ is generated from the noisy audio input.
\vspace{-0.6em}
\item
A separation mask $F$ is constructed using the threshold $\tau(f)$:
For each $(t,f)$ in the spectrogram, we compute:
\vspace{-0.3em}
\begin{equation}
F(t,f) =
\left\{
	\begin{array}{ll}
		1  & S(t,f) > \tau(f) \\
		0 & otherwise
	\end{array}
\right.
\end{equation}
\vspace{-1.6em}
\item
The noisy mel-scale spectrogram $C$ is filtered by the following operation: $P = C \odot F$, where $\odot$ denotes element-wise multiplication.
\vspace{-0.4em}
\item
The clean speech is reconstructed from the predicted mel-scale spectrogram $P$, preserving the original phase of each isolated frequency.
\end{enumerate}

\section{Experiments}
\label{experiments}

\subsection{Datasets}
\label{dataset}


\paragraph*{GRID Corpus}
We performed experiments on the GRID audio-visual sentence corpus \cite{gridcorpus}, a large dataset of audio and video (facial) recordings of 1,000 3-second sentences spoken by 34 people. A total of 51 different words are contained in the GRID corpus.

\vspace{-1em}
\paragraph*{TCD-TIMIT}
We conducted additional experiments on the TCD-TIMIT dataset \cite{harte2015tcd}. This dataset consists of 60 speakers with around 200 videos each, as well as three lipspeakers, people specially trained to speak in a way that helps lipreaders understand their visual speech. The speakers are recorded saying various sentences from the TIMIT dataset \cite{garofolo1993darpa} using both front-facing and $30$ degree cameras.

\vspace{-1em}
\paragraph*{Mixing protocol}
For each one of the experiments, we synthesize audio mixtures from the speech signals of two speakers of the same gender. Given audio signals $s_1(t), s_2(t)$ their mixture is synthesized to be $s_1(t) + s_2(t)$, using the original, unnormalized gain of each source. The signals in all experiments are taken from data unseen when training the relevant \emph{vid2speech} models.

\subsection{Performance evaluation}
\label{ssec:evaluation}
The results of our experiments are evaluated using objective source separation evaluation scores, including SDR, SIR and SAR \cite{vincent2006performance} and PESQ \cite{pesq}. In addition to these measurements, we assessed the intelligibility and quality of our results qualitatively using informal human listening. We strongly encourage readers to watch and listen to the supplementary video available on our project webpage\footnote{Examples of speech separation and enhancement can be found at \url{http://www.vision.huji.ac.il/speaker-separation}}, demonstrating the effectiveness of our approach.

\subsection{Results}

\begin{figure}[!tbh]
\centering

\begin{subfigure}{0.45\textwidth}
\centering
\includegraphics[width=0.45\linewidth]{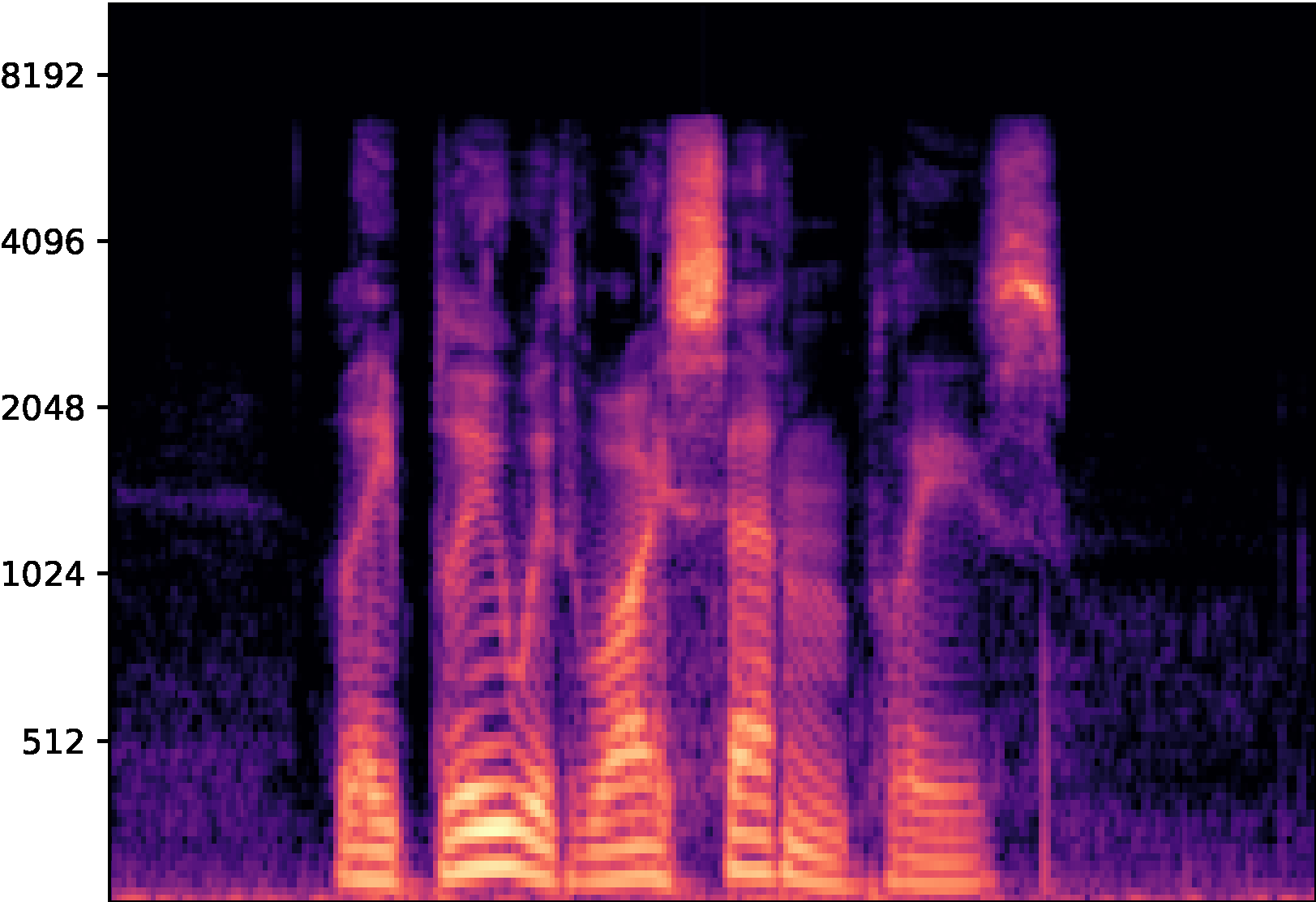}
\includegraphics[width=0.45\linewidth]{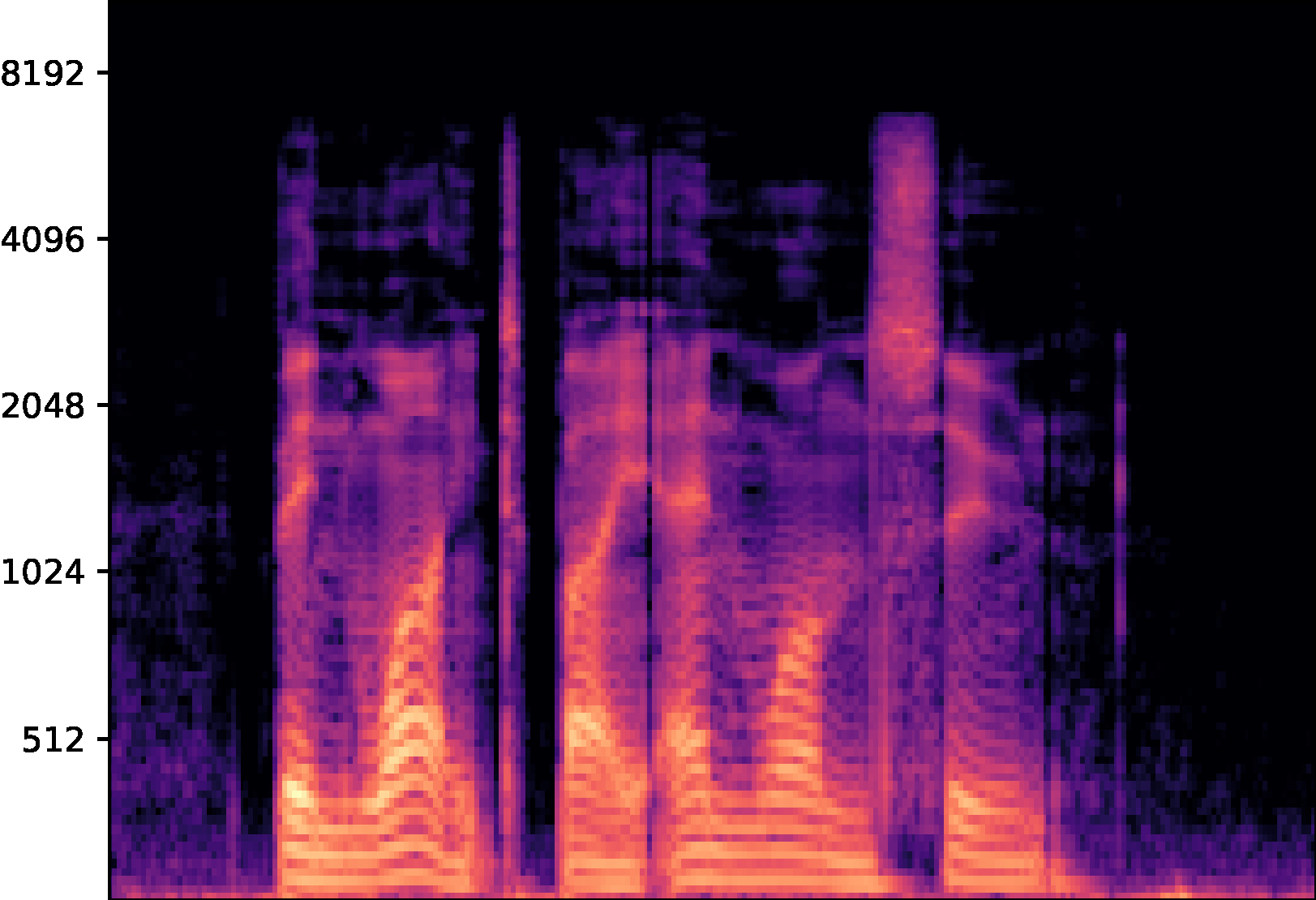}
\caption{Spectrograms of two source voices.}
\end{subfigure}

\begin{subfigure}{0.45\textwidth}
\centering
\includegraphics[width=0.45\linewidth]{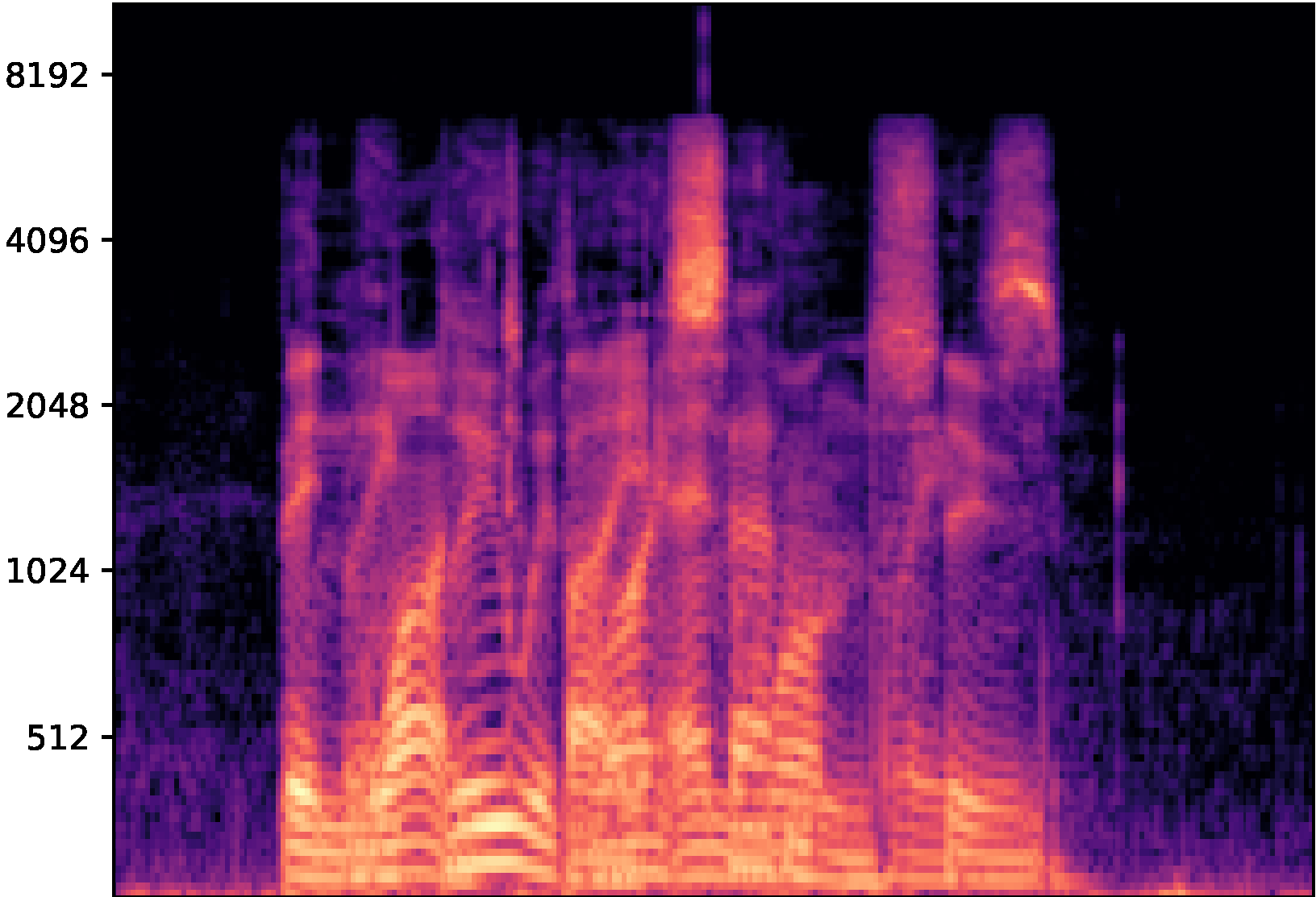}
\caption{Spectrogram of the mixed voices.}
\end{subfigure}

\begin{subfigure}{0.45\textwidth}
\centering
\includegraphics[width=0.45\linewidth]{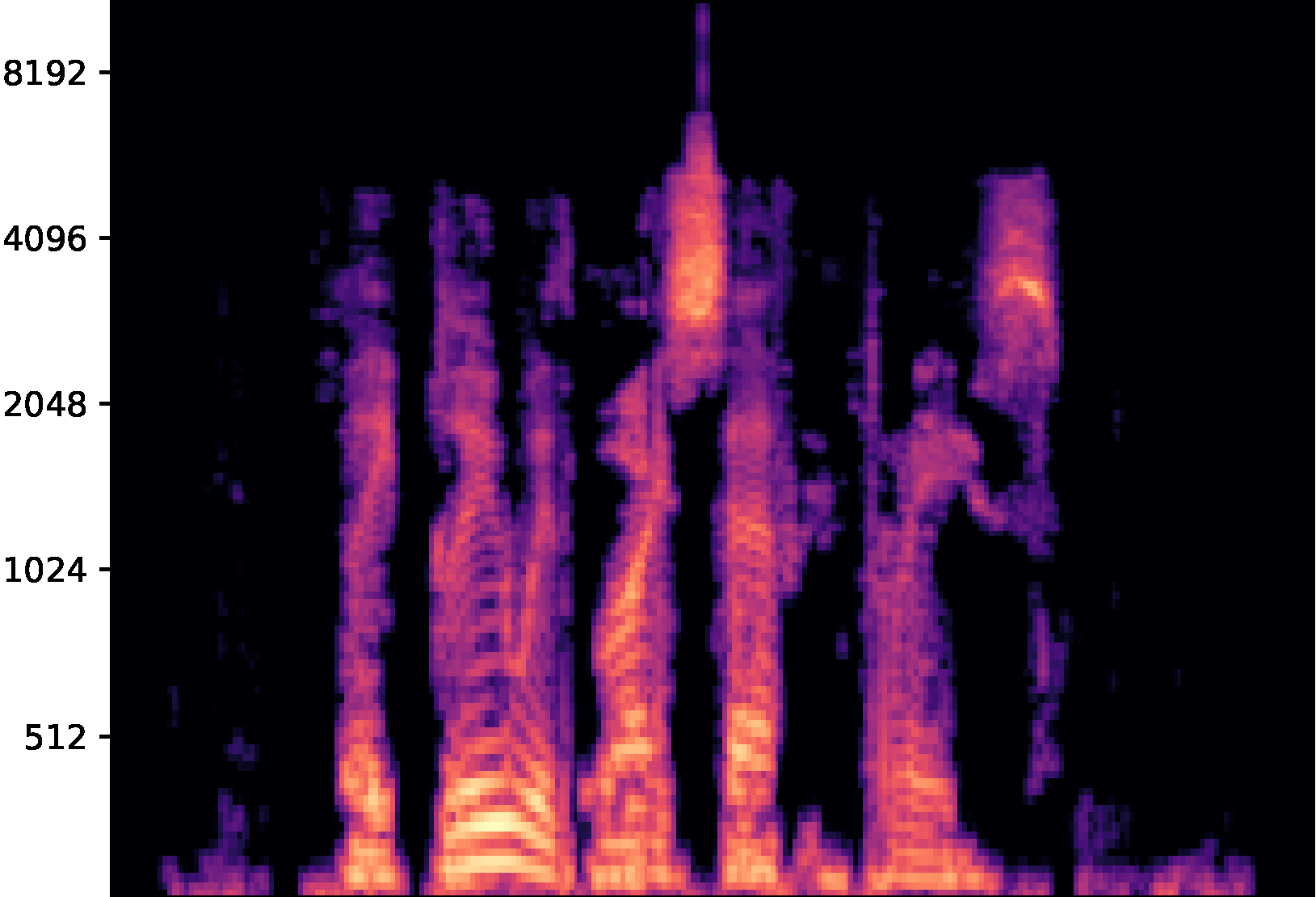}
\includegraphics[width=0.45\linewidth]{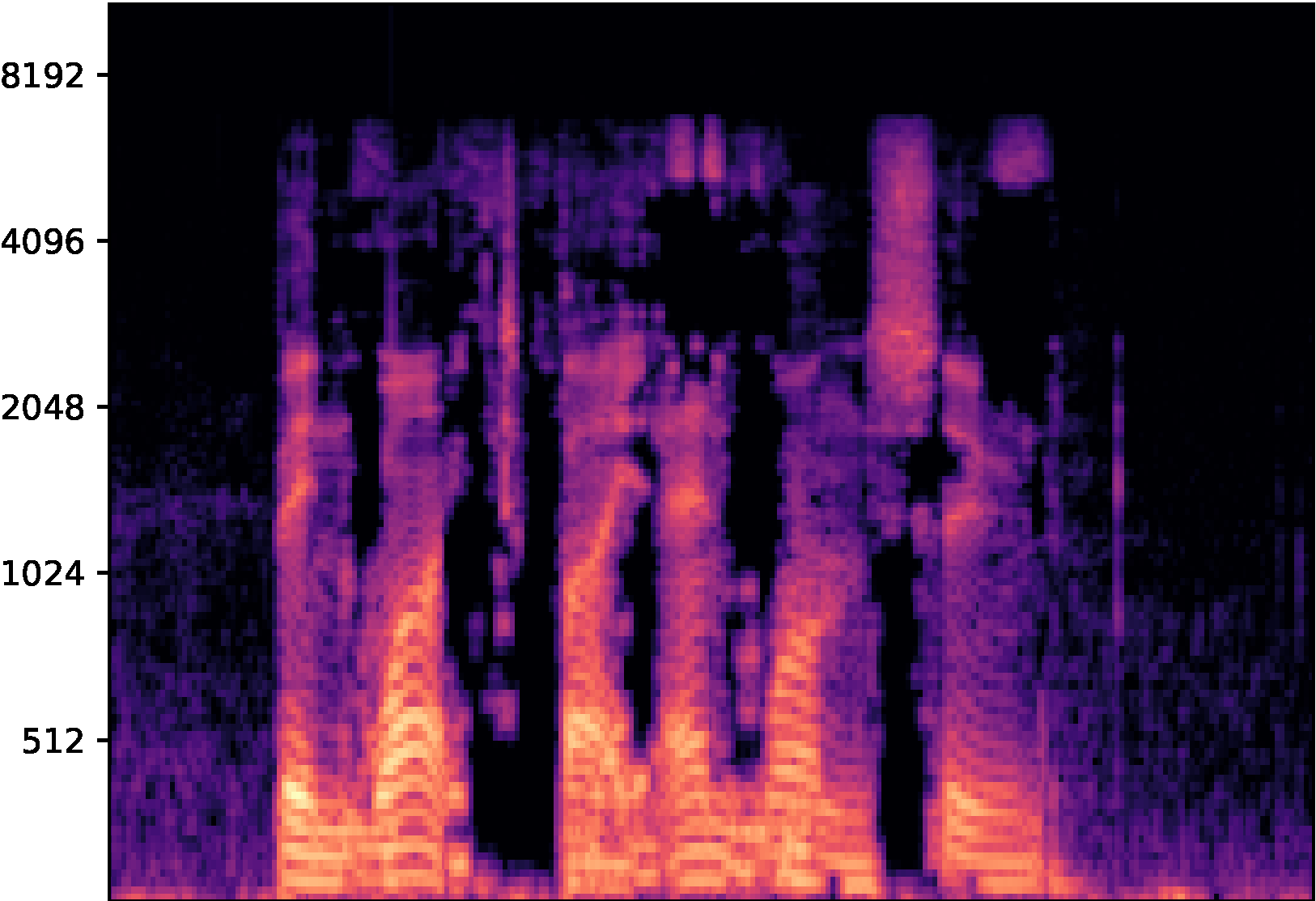}
\caption{Spectrograms of the separated voices.}
\end{subfigure}

\caption{Spectrograms for one segment in our separation testing data from the GRID dataset.}
\label{fig:spectrograms}
\vspace{-0.8em}
\end{figure}

\paragraph*{Separation}
Table \ref{tb:separation_results} shows the results of the separation experiments on synthesized mixtures of sentences spoken from the GRID and TCD-TIMIT datasets. The GRID experiment involved testing on random speech mixtures from two male speakers (\emph{S2} and \emph{S3}). The TCD-TIMIT experiment involved random speech mixtures of a female speaker (\emph{lipspeaker 3}) with her own voice, emphasizing the capabilities of our approach. We present a comparison to results obtained by applying the audio-only method of Huang \emph{et al.}~\cite{huang2014deep}. In addition, we compare to the raw speech predictions generated by \emph{vid2speech}, without applying any of our separation methods. 

It can be seen that the raw speech predictions have reasonable quality (PESQ score) when dealing with a constrained-vocabulary dataset such as GRID. However, \emph{vid2speech} generates low quality and mostly unintelligible speech predictions when dealing with a more complex dataset such as TCD-TIMIT, which contains sentences from a larger vocabulary. In this case, our separation methods have real impact, and the final speech signals sound much better than the raw speech predictions.
We use the spectrograms of ground truth source signals to construct the ideal binary and ratio masks, and present their separation scores as a performance ceiling of our separation method. Examples of the separated spectrograms are shown in Figure \ref{fig:spectrograms}.

\vspace{-1em}
\paragraph*{Enhancement}
Table \ref{tb:enhancement_results} shows the results of enhancement experiments on synthesized mixtures of sentences spoken from the GRID and TCD-TIMIT datasets. The GRID experiment involved random speech mixtures of two male speakers (\emph{S2} as target speaker and \emph{S3} as background speaker). The TCD-TIMIT experiment involved random speech mixtures of two female speakers (\emph{lipspeaker 3} as target and \emph{lipspeaker 2} as background). Here we also present a comparison to the raw speech predictions generated by \emph{vid2speech}. We use the spectrograms of ground truth source signals as an `oracle' to evaluate an upper bound for the performance of our method.

We also evaluated our enhancement method qualitatively on mixtures of speech and non-speech background noise, examples of which can been seen on our project webpage.


\begin{table}[tb]
\centering
\begin{tabular}{lcccc}
\toprule[1.5pt]
\bf  & \bf SDR & \bf SIR & \bf SAR & \bf PESQ\\
\midrule
\bf GRID &						\\
\midrule
Noisy 					& 0.04 & 0.05  & 40.6 & 2.1 \\
\midrule
\emph{Vid2speech} \cite{ephrat2017improved}	& -15.19 & 7.41 & -14.2 & 1.91 \\
Audio-only \cite{huang2014deep} & 1.74	& 2.75	& 6.59	& 1.85 \\
Ours - binary mask		& 5.1 & \textbf{13.02} & 6.41 & 2.07 \\
Ours - ratio mask		& \textbf{5.62} & 8.83 & \textbf{9.49} & \textbf{2.6} \\
\midrule
Ideal binary mask	& 10.6 & 22.03 & 11.03 & 2.9 \\
Ideal ratio mask	& 10.1 & 14.15 & 12.65 & 3.58 \\
\midrule
\midrule
\bf TCD-TIMIT &	\\
\midrule
Noisy 					& 0.15 & 0.15 & 237.17 & 2.26 \\
\midrule
\emph{Vid2speech} \cite{ephrat2017improved}	& -12.99 & 13.53 & -12.26 & 1.41 \\
Audio-only \cite{huang2014deep} & 2.91	& 4.62	& 9.04	& 2.16 \\
Ours - binary mask		& 8.11 & \textbf{17.8} & 9.01 & 2.4 \\
Ours - ratio mask		& \textbf{8.68} & 13.39 & \textbf{11.04} & \textbf{2.71} \\
\midrule
Ideal binary mask	& 15.49 & 28.76 & 15.88 & 3.4 \\
Ideal ratio mask	& 15.19 & 21.61 & 16.6 & 3.86 \\
\bottomrule[1.5pt]
\end{tabular}
\caption{Comparison of the separation quality on the GRID and TCD-TIMIT datasets using binary and ratio masking, along with a comparison to the audio-only separation method of Huang \emph{et al.} \cite{huang2014deep} and raw \emph{vid2speech} \cite{ephrat2017improved} predictions.}
\label{tb:separation_results}
\vspace{-1em}
\end{table}


\begin{table}[h]
\centering
\begin{tabular}{lcc}
\toprule[1.5pt]
\bf & \bf SNR & \bf PESQ \\
\midrule
\bf GRID &						\\
\midrule
Noisy	& -0.63 & 1.83 \\
\midrule
\emph{Vid2speech} \cite{ephrat2017improved}	& -2.51 & 1.93 \\
Ours	& \textbf{2.11} & \textbf{1.97} \\
\midrule
Ideal enhancement	& 2.82 & 2.4 \\
\midrule
\midrule
\bf TCD-TIMIT &	\\
\midrule
Noisy				& 0.97 & 2.19 \\
\midrule
\emph{Vid2speech} \cite{ephrat2017improved}	& -11.19 & 1.42 \\
Ours				& \textbf{4.52} & \textbf{2.1} \\
\midrule
Ideal enhancement						& 9.28 & 2.41 \\
\bottomrule[1.5pt]
\end{tabular}
\caption{Evaluation of the enhancement quality using LTSS as the mask thresholding function.}
\label{tb:enhancement_results}
\end{table}

\vspace{-1em}
\paragraph*{Speech separation of unknown speakers}
Attempts to predict speech of an unknown speaker using a model trained on a different speaker usually led to bad results. In this experiment, we attempted to separate the speech of two `unknown' speakers. First, we trained a \emph{vid2speech} network \cite{vid2speech} on the data of a `known' speaker (\emph{S2} from GRID). The training data consisted of randomly selected sentences ($40$ minutes length in total). Before predicting the speech of each one of the `unknown' speakers (\emph{S3} and \emph{S5} from GRID) as required in the separation method, we fine-tuned the network using a small amount of samples of the actual speaker ($5$ minutes length in total).  Then, we applied the speech separation process to the synthesized mixtures of unseen sentences spoken by the unknown speakers. The results are summarized in Table \ref{tb:independent_results}.

\begin{table}[tb]
\centering
\begin{tabular}{lcccc}
\toprule[1.5pt]

\bf  & \bf SDR & \bf SIR & \bf SAR & \bf PESQ\\
\midrule
Noisy	& 0.04 & 0.04 & 36.14 & 2.14 \\
\midrule
\emph{Vid2speech} \cite{ephrat2017improved}	& -16.37 & 6.55 & -15.19 & 1.76 \\
Ours - binary mask		& 1.85 & \textbf{8.61} & 4.06 & 1.74 \\
Ours - ratio mask		& \textbf{3.06} & 5.86 & \textbf{7.9} & \textbf{2.42} \\
\midrule
Ideal binary mask	& 10.07 & 21.7 & 10.5 & 2.99 \\
Ideal ratio mask	&  9.55 & 13.44 & 12.24 & 3.65 \\
\bottomrule[1.5pt]
\end{tabular}
\caption{Comparison of the separation quality of unknown speakers from GRID corpus using transfer learning.}
\label{tb:independent_results}
\vspace{-0.8em}
\end{table}

\section{Concluding remarks}
\label{sec:conclusion}
This work has shown that high-quality single-channel speech separation and enhancement can be performed by exploiting visual information.
Compared to audio-only techniques mentioned in Sec.~\ref{ssec:related}, our method is not affected by the issue of similar speech vocal characteristics as commonly observed in same-gender speech separation, since we gain the disambiguating power of visual information.

The work described in this paper can serve as a basis for several future research directions. These include using a less constrained audio-visual dataset consisting of real-world multi-speaker and noisy recordings. Another interesting point to consider is improving the performance of voice recognition systems using our enhancement methods. Implementing a similar speech enhancement system in an end-to-end manner may be a promising direction as well.

~\\
\noindent
{\bf Acknowledgment.} This research was supported by Israel Science Foundation and by Israel Ministry of Science and Technology.
\bibliographystyle{IEEEbib}
\bibliography{references}

\end{document}